\documentclass[journal]{IEEEtran}
\usepackage{cite}

\ifCLASSINFOpdf
   \usepackage[pdftex]{graphicx}
\else
\fi

\usepackage{url}
\usepackage{bm}
\usepackage{multirow}
\usepackage{tabularx}
\usepackage{color}
\usepackage{hyperref}
\usepackage{caption}
\usepackage{soul}

\usepackage{booktabs}

\usepackage{times}
\usepackage{epsfig}
\usepackage{graphicx}
\usepackage{amsmath}
\usepackage{amssymb}
\usepackage{comment}
\usepackage{authblk}
\usepackage{multirow}
\usepackage{tabularx}
\usepackage{balance}
\usepackage{subfigure}
\usepackage{balance}
\usepackage{algorithm}
\usepackage[table]{xcolor}
\usepackage[noend]{algpseudocode}
\usepackage{cite}
\usepackage{enumitem}
\setlist{nosep}

\DeclareRobustCommand{\hlcyan}[1]{{\sethlcolor{white}\hl{#1}}}

\begin{document}
%
\title{MirrorNet: Bio-Inspired Camouflaged Object Segmentation}
%
%
%

\author[1]{Jinnan Yan}
\author[2]{Trung-Nghia Le}
\author[3, 4]{Khanh-Duy Nguyen}
\author[3, 6, 7]{Minh-Triet Tran}
\author[5]{Thanh-Toan Do}
\author[1]{Tam V. Nguyen*\thanks{*Corresponding author. {\it e-mail: tamnguyen@udayton.com}}}

\affil[1]{University of Dayton, US}
\affil[2]{National Institute of Informatics, Japan}
\affil[3]{Vietnam National University, Ho Chi Minh City, Vietnam}
\affil[4]{University of Information Technology, VNU-HCM, Vietnam}
\affil[5]{Monash University, Australia}
\affil[6]{University of Science, Ho Chi Minh City, Vietnam}
\affil[7]{John von Neumann Institute, VNU-HCM, Vietnam}

\markboth{IEEE Access}%
{MirrorNet: Bio-Inspired Camouflaged Object Segmentation}
%



\maketitle

\begin{abstract}
   Camouflaged objects are generally difficult to be detected in their natural environment even for human beings. In this paper, we propose a novel bio-inspired network, named the MirrorNet, that leverages both instance segmentation and mirror stream for the camouflaged object segmentation. Differently from existing networks for segmentation, our proposed network possesses two segmentation streams: the main stream and the mirror stream corresponding with the original image and its flipped image, respectively. The output from the mirror stream is then fused into the main stream's result for the final camouflage map to boost up the segmentation accuracy. Extensive experiments conducted on the public CAMO dataset demonstrate the effectiveness of our proposed network. Our proposed method achieves $89\%$ in accuracy, outperforming the state-of-the-arts.
\end{abstract}


%
\IEEEpeerreviewmaketitle

\section{Introduction}



\begin{figure}[!t]
	\centering
	\includegraphics[width=1\linewidth]{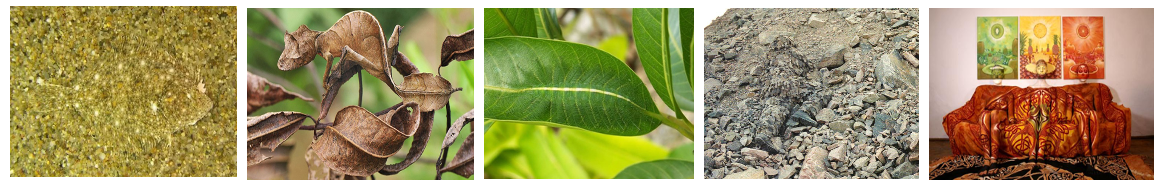}
	\caption{\hlcyan{A few examples of camouflaged objects in CAMO dataset}~\cite{CAMO}. \hlcyan{From left to right, naturally camouflaged objects (e.g., fish, chameleon, insect) are followed by artificially camouflaged objects (e.g., soldier, body painting).}} 
	\label{fig:example}
\end{figure}

The term ``camouflage'' was originally used to describe the behavior of \hlcyan{an animal or insect trying to hide itself from its surroundings to hunt or avoid being hunted}~\cite{Sujit-ICEECS2013}, \hlcyan{namely \textit{naturally camouflaged objects}}~\cite{CAMO}. \hlcyan{This ability is useful to reduce risk of being detected and increase their survival probability. For example, chameleons or fishes can change appearance of their bodies to match color and pattern of surrounding environments} (see Fig.~\ref{fig:example}). \hlcyan{Human adopted this mechanism and began to apply it widely on the battlefield. For example, soldiers and war equipment are applied the camouflage effect by dressing or coloring their appearance to blend them with their surroundings} (see Fig.~\ref{fig:example}), \hlcyan{namely \textit{artificially camouflaged objects}}~\cite{CAMO}. \hlcyan{Artificial camouflage has been also applied into entertainment (e.g., magic show) or art (e.g., body painting). Figure} \ref{fig:example} \hlcyan{shows a few examples of both naturally and artificially camouflaged objects in real life, in which these camouflaged objects are not identified obviously.}

Autonomously detecting/segmenting camouflaged objects is thus a difficult task where discriminative features do not play an important role since we have to ignore objects that capture our attention. While detecting camouflaged objects is technically challenging on the one hand, it is beneficial in various practical scenarios, on the other, to include surveillance systems and search-and-rescue missions. 



\begin{figure*}[!t]
	\centering
	\includegraphics[width=1\textwidth]{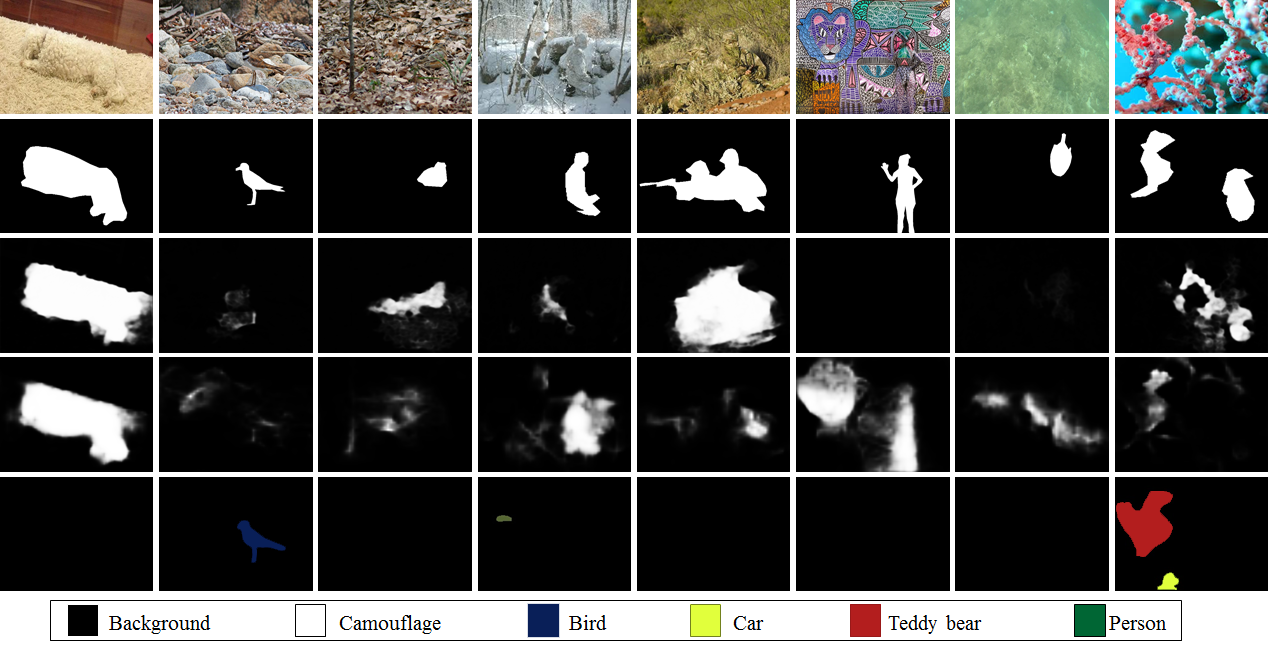}
	\caption{Failure examples of the state-of-the-art method on camouflaged object segmentation. From top to bottom: the original images, the ground-truth maps, the camouflage maps of ANet-DHS~\cite{CAMO}, and ANet-SRM~\cite{CAMO}, and the instance segmentation results of Mask R-CNN~\cite{Kaiming-ICCV2017}. We remark that all these methods are fine-tuned on the camouflage dataset~\cite{CAMO}.
	} 
	\label{fig:failure}
\end{figure*}





Figure~\ref{fig:failure} shows several examples that camouflaged objects are failed to be detected by state-of-the-art object segmentation networks. Moreover, there are plenty of creatures in nature that have evolved over time to confuse themselves with their surroundings and even harder to be detected. Therefore, the study of detecting these less obvious objects which we call it camouflaged object in this paper, is necessary in the field which targets detecting all objects in all scenes. Note that camouflage is a very subjective concept here. An object, such as a person which is regarded as a common class in MS-COCO~\cite{Lin-ECCV2014}, can be considered as a camouflaged object when he is hiding himself as a sniper. In other words, we treat all objects that are too similar to their surroundings because of their color, texture, or both as camouflaged object and its complement set as non-camouflaged object. It is time-consuming and laborious to collect data from all target objects in particular scenes which presents camouflage to improve models, and it is not well adaptable. Moreover, what camouflage has in common is that its features are very little different from its surroundings. Therefore, it is reasonable to treat all objects of different classes but similar to the background as one class.

Most state-of-the-art Convolutional Neural Network (CNN)-based models~\cite{resnext, Szegedy-AAAI2017, Huang-CVPR2017} simulate human brain. Therefore, the CNN-based models may also be fooled just like human~\cite{Alcorn_2019_CVPR}. This is also true for camouflaged objects, which attempt to fool our human visual perception. Through time, they evolve to well blend their appearance to the background. On the other hand, color-blind people are often better at detecting camouflaged objects~\cite{colorblind}. This is perhaps because they rely less on colors and more on form and texture to discern the world around them. This motivates us to look at other bio-inspired features. 

An object becomes a successful camouflaged object when it elegantly blends into the surrounding environment to create a familiar natural scene that can hide the object. By changing the viewpoint of the same scene, we expect to have a possibility to escape from such an illusion. This visual-psychological phenomenon motivates our bio-inspired solution to detect camouflaged objects by changing the viewpoint of the scene. We realize that just a simple flipping operation also can generate new views of the same scene. Indeed, the flipped images accidentally break the natural layout which leads to the much difference between the background and the camouflaged objects. 

Inspired by above visual-psychological phenomenon, we are looking at a method to break the capability to prevent human/machine vision from recognizing camouflaged objects through changing the viewpoint. In this paper, we propose a new simple yet efficient bio-inspired framework, which greatly improves the existing camouflage object recognition performance. The framework leverages the advantages of cutting-edge models~\cite{He-CVPR2016,resnext} , which are well trained on large-scale datasets like ImageNet~\cite{Russakovsky-IJCV2015} and MS-COCO~\cite{Lin-ECCV2014} datasets, making the performance of this framework surpass ANet~\cite{CAMO} which are the state-of-the-art camouflaged object segmentation. The proposed framework consists of two streams, the main stream for segmenting original image and the mirror stream utilizing flipped image, whereas the flipped image stream exploits the bio-inspired effort to break the natural settings of camouflaged objects. Indeed, viewpoint change by flipping operation is useful to escape from the easy-to-be-fooled appearance of camouflaged objects. Leveraging this simple but effective approach can detect images’ mapping to discover more details, to further improve the performance. To solve the insufficiency of limited training data, we also utilize object-level flipping for natural data augmentation for network training. Extensive experiments on the benchmark CAMO dataset~\cite{CAMO} for camouflaged object segmentation show the superiority of our proposed method over the state-of-the-arts. The code and results will be published on our website\footnote{\url{https://sites.google.com/view/ltnghia/research/camo}}\textsuperscript{,}\footnote{\url{https://sites.google.com/site/vantam/research}}.



Our contributions are as follows. 





\begin{itemize}
	\item Differently from the state-of-the-art ANet~\cite{CAMO}, where classification stream and segmentation stream are applied on the whole image, in this work, we  first obtain object proposal, then we apply segmentation on each object proposal.
	\item As an effort to break the natural setting of camouflaged objects, we integrate additional mirror stream which utilizes horizontally flipped images, for mirror detection to support the main stream. We introduce the Data Augmentation in the Wild to solve the data insufficiency.
	\item Last but not least, we conduct extensive experiments to demonstrate the performance of the proposed framework. 
\end{itemize}

The rest of this paper is organized as follows. Section~\ref{sec:related} first reviews related work. Section~\ref{sec:framework}  briefly introduces the motivation and discusses details the proposed framework. Section~\ref{sec:results} reports the experimental results and the ablation study. Finally, Section~\ref{sec:conclusion} draws the conclusion and paves way to the future work.

\section{Related Work} 
\label{sec:related}


In literature, the mainstreams of the computer vision community mainly focus on the detection/segmentation of the non-camouflaged objects, \textit{i.e.}, salient object, objects with predefined classes. There exist many  research works on both general object detection~\cite{Girshick-CVPR2014, fast, Ren-NIPS2015, ltnghia-IV2020}, and salient object segmentation~\cite{Wang-ICCV2017, egnet, cpd, pfanet, basnet,  ltnghia-WACV2019, AH, ltnghia-PSIVT2015, SP, ltnghia-BMVC2017, ltnghia-TIP2018}. Meanwhile, the camouflage object recognition has not been  well explored in the literature. Early works related to camouflage are dedicated to detecting the foreground region even when some of their texture is similar to the background~\cite{Galun-ICCV2003, Song-ICMT2010, Xue-MTA2016}. These works distinguish foreground and background based on simple features such as color, intensity, shape, orientation, and edge. A few methods based on handcrafted low-level features (\textit{i.e.}, texture~\cite{Pan-MAS2011, Liu-TIP2012, Sengottuvelan-ICETET2008} and motion~\cite{Yin-PE2011, Gallego-ICIP2014}) are proposed to tackle the problem of camouflage detection. However, all these methods work for only a few cases of simple and non-uniform background, thus their performances are unsatisfactory in camouflaged object segmentation due to strong similarity between the foreground and the background. Recently, Le \textit{et al.}~\cite{CAMO} proposed an end-to-end network, called ANet, for camouflaged object segmentation through integrating classification information into segmentation. Fan \textit{et al.}~\cite{sinet} introduced SINet with two main modules, namely the search module \textit{S} and the identification module \textit{I}. Recently, Le \textit{et al.}~\cite{camo_aaai} explored the camouflaged instance segmentation by training Mask RCNN~\cite{Kaiming-ICCV2017} on CAMO dataset~\cite{CAMO}.

Differently from existing work, which only interferes the training process, in this work, we also utilize \textit{bio-inspired mirror-stream} in the inference process to guide the network towards the right track for camouflage object segmentation. We remark that our mirror-stream and image flipping augmentation of existing work are different for the purpose of usage and philosophy. For data augmentation, it applies image flipping in only the training phase to improve the performance of networks. It means that they consider the original image and its flipped image as two totally different images. They want to learn as much as possible cases that can appear. Meanwhile, our flip-stream is independent from image flipping augmentation. We apply flipped images together with the original images in the inference phase. It means that we consider the normal image and the flipped image as two-faces of the original image. Besides solving the standard case (normal image), we also want to solve the rare case (flipped image).

\section{Proposed Framework}
\label{sec:framework}
\subsection{Bio-inspired Motivation}
\label{sec:motivation}
In biological vision studies, there exist viewpoint-invariant theories and viewpoint-dependent theories~\cite{wilson2003does,burghund2000viewpoint,li2009computational}. In viewpoint-invariant theories, once a particular object has been stored, the recognition of that object from novel viewpoints should be unaffected, provided that the necessary features can be recovered from that view. In view-dependent theories, once a particular object has been stored,
recognition of that object from novel views may be impaired, relative to recognition of previously stored views. However, these theories were formed from the simple setting with the outstanding stimuli (non-camouflaged objects) and the clear background, \textit{i.e.}, white background. Indeed, it is natural for human vision to easily detect non-camouflaged objects such as salient objects since they are outstanding from the background, while it is harder for human to detect camouflaged objects since they are similar to the background. In other words, the visual difference between the background and the non-camouflaged object(s) should be larger than the one between the background and the camouflaged object(s). However, when we flip the images, humans are not fooled by the natural settings and notice differences in the images. Therefore, the flipped images accidentally break the natural layout which leads to the larger distance between the background and the camouflaged objects.


To prove this conclusion intuitively, we compute \textbf{the visual difference between the main objects (the camouflaged and the non-camouflaged objects) with the background}. In particular, we adopt the well-known deep learning model FCN~\cite{Long-ICCV2015} and its variants to compute the difference distance. We use the pretrained model on PASCAL-VOC~\cite{Everingham-ICCV2010} (with 20 semantic classes and the background class). We extract the confidence score at the second last layer of FCN ($h \times w \times 21$). Then, we compute the mean semantic vectors $s_{bg}$, $s_{fg}$ for the background and the camouflaged/non-camouflaged object regions (obtained from the ground truth maps), respectively. This ends up with 21-dim vector for both regions. $\ell_2$ normalization is applied for both $s_{bg}$ and $s_{fg}$. Then the visual difference $d$ between the main objects (the camouflaged/non-camouflaged objects) with the background is computed as: $d = \|s_{fg} - s_{bg}\|_2$, where $\|.\|_2$ is the Euclidean distance. We also compute the difference between the camo/non-camo object with the background in terms of color (RGB and $l\alpha\beta$), and texture (texton~\cite{texton}).

\begin{table}[!t]
	\small	
	\centering
	\begin{tabular}{|l|c|c|c|}
		\hline
		\textbf{Method} & \textbf{Non-Camo} & \textbf{Camo} & \textbf{Camo-Flipped }\\
		\hline
		\textbf{Color (RGB)} & 0.112 & 0.085 & 0.085  \\
		
		\hline
		\textbf{Color ($l\alpha\beta$)} & 0.217 & 0.143 & 0.143  \\
		
		\hline
		
		\textbf{Texton~\cite{texton} } & 0.323 & 0.195 & 0.195  \\
		
		\hline
		\textbf{FCN (32s)~\cite{Long-ICCV2015}} & 0.520 & 0.352 & 0.354  \\
		\hline
		\textbf{FCN (8s)~\cite{Long-ICCV2015}} & 0.641 & 0.409 & 0.412  \\
		\hline
		\textbf{CRF-RNN~\cite{CRF}} & 0.786 & 0.460 & 0.462  \\
		\hline
	\end{tabular}
	\caption{The visual difference between the camouflaged/non-camouflaged objects and the background with different methods. 
	} 
	\label{table:motivation}
\end{table}

\begin{figure*}[t]
	\centering
	\includegraphics[width=1\textwidth]{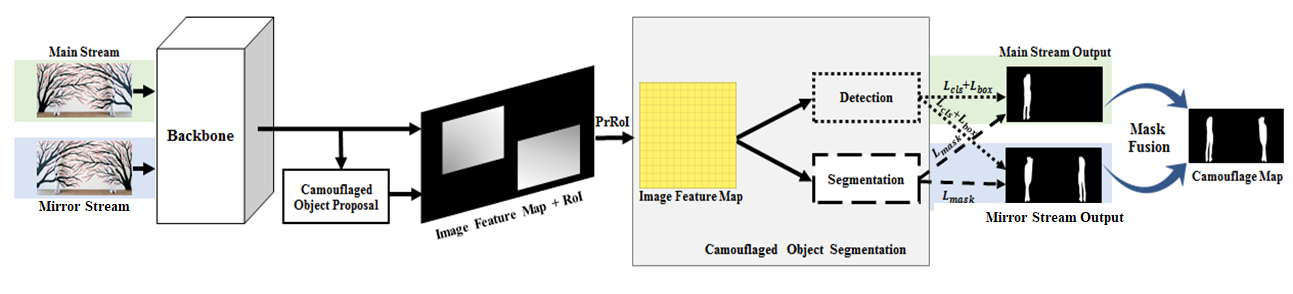}
	\caption{The overview of our proposed framework. MirrorNet consists of two streams, namely, the main stream for original image segmentation and the mirror stream for horizontally flipped image segmentation.}
	\label{fig:framework}
\end{figure*}

Table~\ref{table:motivation} shows the visual difference with different  settings. The difference in terms of color is much smaller than the ones of features extracted from deep learning models. $l\alpha\beta$ shows a larger distance than RGB in the color space. \hlcyan{Meanwhile, the textural features, i.e.,  Texton}~\cite{texton}\hlcyan{, yield the larger distance which indicates the usefulness of texture in the task of  camouflaged object segmentation. The even larger distances in feature space show that features from deep learning models may be helpful to detect and segment camouflaged objects. Note that the features extracted from FCN, a CNN-based model, actually capture the information of textures and edges. In fact, FCN and CNN networks contain convolutional layers which learn spatial hierarchies of patterns by preserving spatial relationships. In these networks, the first convolutional layer can learn basic elements such as texture and edges}~\cite{QuocLe}\hlcyan{. Then, the second convolutional layer can learn patterns composed of basic elements learned in the previous layer. The training process continues until the deep network learns  significantly complex patterns and abstract visual concepts. This demonstrates the importance of low-level information such as texture and edges in the task of camouflaged object segmentation.} 

In addition, the difference values of camo and camo-flip with the background are identical in terms of color. It shows that the flipped stream is not useful in terms of color. However, when we look at the difference in the feature space, the camo and non-camo values are different (non-camo value is slightly better). It shows that the camo flip in the feature space may be helpful for the task. 

As a closer look, the distance between main object and background is larger in non-camouflaged and smaller in camouflaged images in all FCN and CRF-RNN settings. In camouflaged images, most pixels of camouflaged objects are classified as \textit{background class} which leads to the small distance. Meanwhile, most pixels of the main non-camouflaged objects are classified with \textit{non-background classes} which cause the larger distance. We are also interested in the horizontally flipped images. Therefore, we also compute the visual difference between the background and the camouflaged objects in the horizontally flipped images. As shown in Table~\ref{table:motivation}, the camouflaged-flipped images generally produce larger distance. It seems that the horizontally flipped images may contain useful information for the segmentation task. Note that the images of camouflaged objects are taken with the natural settings. Therefore, \textit{the flipped images accidentally break the natural layout} which lead to the larger distance between the background and the camouflaged objects. This one absolutely paves way to our proposed method.

The camouflaged object contains \textit{intrinsic} and \textit{extrinsic} information. The intrinsic information means the difference between the camouflaged object and the background. The small distance may be observed in the color space. The distance may be larger in another space, \textit{i.e.}, deep learned feature space, semantic space. Meanwhile, the extrinsic information means the external change onto the camouflaged object, for example, rotating, translating, and flipping, so that the camouflaged object can be better recognized. Therefore, our proposed framework captures both intrinsic and extrinsic information of the camouflaged object.

%


\subsection{MirrorNet Overview}

Figure \ref{fig:framework} depicts the overview of our proposed MirrorNet, the bio-inspired network for camouflaged object segmentation. The MirrorNet actually consists of two streams, the main stream for original image segmentation and the mirror stream for flipped image segmentation, whereas the horizontally flipped image stream exploits the bio-inspired  effort to break the natural setting of camouflaged objects. Each stream traverses through the camouflaged object proposal, camouflaged object segmentation and yields the segmentation masks. The two output masks from the two streams are finally fused to produce a pixel-wisely accurate camouflage map which uniformly covers camouflaged objects. Here, the camouflaged object segmentation targets at the \textit{intrinsic} information. Meanwhile, the mirror stream aims to capture the \textit{extrinsic} information.

\subsection{Network Design and Architecture}

\begin{figure*}[!t]
	\centering
	\includegraphics[width=\linewidth]{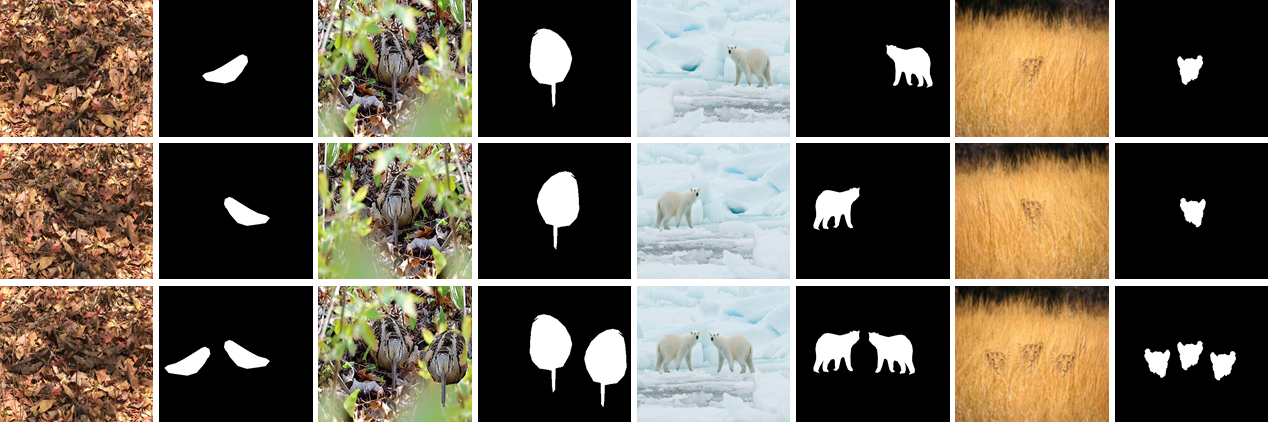}
	\caption{Exemplary samples of Data Augmentation in the Wild. From top to bottom: the original images with the ground-truth map, the flipped images, the cloning instance images.}
	\label{fig:aw}
\end{figure*}


\subsubsection{Camouflaged Object Proposal} 
MirrorNet first attempts to localize the possible positions containing the camouflaged objects via Camouflaged Object Proposal component. Inspired by~\cite{Ren-NIPS2015}, we aim to find the object positions, the object classes, and their camouflage masks in images simultaneously. Follow the standard design in computer vision, the object position is defined by a rectangle with respect to the top-left corner of the image; the object class is defined over the rectangle; the object mask is encoded at every pixel inside the rectangle. Ideally, we want to detect all relevant objects in the image and map each pixel in these objects to its most probable camouflage/non-camouflage label. 

Here, our Camouflaged Object Proposal component adopts the Region Proposal Network (RPN)~\cite{Ren-NIPS2015}. RPN shares weights with the main convolutional backbone and outputs bounding boxes (RoI / object proposal) at various sizes. For each RoI, a fixed-size feature map 
is pooled from the image feature map using the RoIPool layer~\cite{Ren-NIPS2015}. The RoIPool layer works by dividing the RoI into a regular grid and then max-pooling the feature map values in each grid cell. This quantization, however, causes misalignments between the RoI and the extracted features due to the harsh rounding operations when mapping the RoI coordinates from the input image space to the image feature map space and when dividing the RoI into grid cells. In order to address this problem, our Camouflaged Object Proposal component integrates Precise RoI Pooling (PrRoI)~\cite{PrROI}. Indeed PrRoI Pooling uses average pooling instead of max pooling for each bin and has a continuous gradient on bounding box coordinates. That is, one can take the derivatives of some loss function with respect to the coordinates of each RoI and optimize the RoI coordinates. PrRoI Pooling is also different from the RoI Align proposed in Mask R-CNN. PrRoI Pooling uses a full integration-based average pooling instead of sampling a constant number of points. This makes the gradient with respect to the coordinates continuous.

\subsubsection{Camouflaged Object Segmentation}
Following~\cite{Ren-NIPS2015}, we use a multi-task loss $\mathcal{L}$ to jointly train the bounding box class, the bounding box position, and the camouflage map (mask) as follows:
\begin{equation}
\mathcal{L} = \mathcal{L}_{cls} + \mathcal{L}_{loc} + \mathcal{L}_{mask},
\end{equation}
where $\mathcal{L}_{cls}$ and $\mathcal{L}_{loc}$ are the output of the detection branch. Meanwhile, $\mathcal{L}_{mask}$
is defined on the output of the segmentation branch. The object classification loss $\mathcal{L}_{cls}(p, u)$ is the multinomial cross entropy loss computed as follows:
\begin{equation}
\mathcal{L}_{cls}(p, u) = -\log{p_u},
\end{equation}
where ${p_u}$ is the softmax output for the true class $u$. The bounding box regression loss $\mathcal{L}_{loc}(t^u, v)$ is computed as the Smooth L1 loss~\cite{fast} between the regressed box offset $t^u$ (corresponding to the ground-truth object class $u$) and the ground-truth box offset $v$:

\begin{equation}
\mathcal{L}_{loc}(t^u, v) = \sum_{i \in \{x,y,w,h\}} Smooth_{L1}(t^u_i-v_i)
\end{equation}

The segmentation loss $\mathcal{L}_{mask}(m, s)$ is the multinomial cross entropy loss computed as follows:

\begin{equation}
\mathcal{L}_{mask}(m,s) = \frac{-1}{N}\sum_{i \in RoI}{\log{m^i_{s_i}}},
\end{equation}
where $m^i_{s_i}$ is the softmax output at pixel $i$ for the true label $s_i$, $N$ is the number of pixels in the RoI.

\subsubsection{Mask Fusion} 

We first flip all detected bounding boxes and their corresponding camouflage map (mask) from the mirror stream. Then, we use threshold $\theta=0.5$ to eliminate bounding boxes with low prediction scores. After that, the prediction scores of bounding boxes from the two streams are sorted in descending order. Then, we apply the ``winner take all'' strategy to prune the redundant boxes. In particular, for each bounding box from highest prediction scores to lowest scores, we check the  bounding boxes with lower scores, if the lower score bounding box and the higher score bounding box have 50\% mutual overlap, the bounding box with lower score will be excluded. Then, we discard the bounding boxes (and their corresponding camouflage maps) classified as non-camouflage.  


Finally we accumulate the camouflage maps (masks) from the retaining bounding boxes and then normalize the output, resulting the final camouflage map.

\subsection{Data Augmentation in the Wild}

Since non-camouflaged object segmentation attracts most attention compared to camouflaged object segmentation, there are only a few relevant datasets, and most of them have the problem of too few samples. Therefore, we adopt the recently built CAMO dataset~\cite{CAMO} which proposed and benchmarked by the previous state-of-the-art camouflaged object segmentation, for the training of instance segmentation framework. The dataset is divided into camouflage and non-camouflage categories, each containing 1,000 training and 250 test set, and a total of 2,500 manually annotated ground truths. Most of the dataset images are mammals, insects, birds, and aquatic animals, each with approximately similar proportions, and a small number of reptiles, human art, soldiers, and amphibians. The diversity of species in this dataset makes our model very adaptable, but it must be pointed out that it also has insufficient samples compared to mainstream datasets like COCO~\cite{Lin-ECCV2014}.

We first compute the number of connected components on the binary ground truth maps of CAMO dataset (the training part). There are many scenarios where multiple connected components belong to one instance. To avoid cherry pick the training samples, we exclude all training images with more than 2 connected components. Then, we compute the bounding box for the single component in the images. In order to increase the number of training samples, we introduce the Augmentation in Wild. In particular, we first perform translating and flipping instances. Furthermore, it is essentially different from data augmentation for non-camouflaged objects, because to determinate whether an object is camouflaged is not only depends on its own features, but also its surroundings. Therefore, we also clone the object instances and place them onto different image regions with a small color difference in the background. Figure~\ref{fig:aw} shows some samples of augmented data. In this way, we increase the number of training samples, thus alleviating the problem of insufficient data.

\subsection{Implementation Details}
To demonstrate the generality and flexibility of our proposed MirrorNet, we employ various recent state-of-the-art backbones, \textit{i.e.},  ResNet~\cite{He-CVPR2016} and ResNeXt~\cite{resnext}. Our implementation is based on the published code of Mask R-CNN \footnote{https://github.com/facebookresearch/maskrcnn-benchmark}, which can be adopted to train camouflaged object proposal and camouflaged object segmentation components. Furthermore, we replace the ROI Align layer with a Precise RoI Pooling layer (PrRoI)~\cite{PrROI}. 

The training process is conducted by fine-tuning the available pre-trained model on camouflage images and non-camouflage images of the CAMO dataset~\cite{CAMO} and our Data Augmentation in the Wild. In particular, we set the size of each mini-batch to 256 and used the Stochastic Gradient Descent (SGD) optimization with a moment $\beta=0.9$ and a weight decay of $0.0001$. We trained our network for 120k iterations with the base learning rate of $0.00125$, which is decreased by $10$ times every time we reach the next steps at 80k iterations and 100k iterations. We also remark that we implemented our method in PyTorch, and conducted all the experiments on a computer with a 2.40GHz processor (Intel Xeon CPU E5-2620), 64 GB of RAM, and one GeForce GTX TITAN X GPU. 


\section{Experiments}
\label{sec:results}
In this section, we first introduce dataset and evaluation criteria used in experiments.  We compare our MirrorNet with state-of-the-art methods on the CAMO dataset~\cite{CAMO}, to demonstrate that the instance segmentation and the mirror stream can boost up the camouflaged object segmentation. We also present the efficiency of our general MirrorNet through the ablation study.

\subsection{Benchmark Dataset and Evaluation Criteria}


We used the entire the testing images in the CAMO dataset for the evaluation. We note that the zero-mask ground-truth labels (all pixels have zero values) are for the non-camouflaged object images. 

Similarly to \cite{CAMO}, we used the F-measure ($F_\beta$)~\cite{Achanta-CVPR2009}, Intersection Over Union (IOU)~\cite{Long-ICCV2015}, and Mean Absolute Error (MAE) as the metrics to evaluate obtained results. The first metric, F-measure, is a balanced measurement between precision and recall as follows: 
\begin{equation}
{F_\beta } = \frac{{\left( {1 + {\beta ^2}} \right)Precision \times Recall}}{{{\beta ^2} \times Precision + Recall}}.
\end{equation}
Note that we set $\beta^2=0.3$ as used in \cite{Achanta-CVPR2009,AH} to put an emphasis on precision. 
IOU is the area ratio of the overlapping against the union between the predicted camouflage map and the ground-truth map. Meanwhile, MAE is the average of the pixel-wise absolute differences between the predicted camouflage map and the ground-truth.

For MAE, we used the raw grayscale camouflage map. For the other metrics, we binarized the results depending on two contexts. In the first context, we assume that camouflaged objects are always present in every image like salient objects; we used an adaptive threshold~\cite{Jia-ICCV2013} $\theta=\mu+\eta$ where $\mu$ and $\eta$ are the mean value and the standard deviation of the map, respectively. In the second context which is much closer to a real-world scenario, we assume that the existence of camouflaged objects is not guaranteed in each image; we used the fixed threshold $\theta=0.5$.

As proposed in~\cite{sinet}, we also use E-measure ($E_{\phi}$)~\cite{ephi},  S-measure ($S_{\alpha}$)~\cite{smeasure}, and weighted F-measure ($F_{\beta}^{\omega}$)~\cite{weightedf} as alternative metrics.

\subsection{Comparison with FCN and ANet Variants} 

\begin{table*}[t]
	\centering
	\caption{Comparison with FCN and ANet variants on two settings: Only Camouflaged Images (the left part), and the full CAMO dataset (the right part). The evaluation is based on F-measure~\cite{Achanta-CVPR2009} (the higher the better), IOU~\cite{Long-ICCV2015} (the higher the better), and MAE (the lower the better). The 1st and 2nd places are shown in \textcolor{blue}{\textbf{blue}} and \textcolor{red}{\textbf{red}}, respectively. 
	}
	\label{tab:result}
	\resizebox{1\textwidth}{!}{
		
		\begin{tabular}{|l|l|c|cc|cc|l|c|cc|cc|}
			\toprule
			& \textbf{Dataset in test} & \multicolumn{5}{c|}{\textbf{Only Camouflaged Images}} &  & \multicolumn{5}{c|}{\textbf{Full CAMO dataset}} \\
			\cline{1-7} \cline{9-13} 
			\textbf{Groups} & \textbf{Method} & \textbf{} & \multicolumn{2}{c|}{\textbf{Adaptive Threshold}} & \multicolumn{2}{c|}{\textbf{Fixed Threshold}} &  & \textbf{} & \multicolumn{2}{c|}{\textbf{Adaptive Threshold}} & \multicolumn{2}{c|}{\textbf{Fixed Threshold}} \\
			& & \textbf{MAE $\Downarrow$} & \textbf{F$_\beta$ $\Uparrow$} & \textbf{IOU $\Uparrow$} & \textbf{F$_\beta$ $\Uparrow$} & \textbf{IOU $\Uparrow$} &  & \textbf{MAE $\Downarrow$} & \textbf{F$_\beta$ $\Uparrow$} & \textbf{IOU $\Uparrow$} & \textbf{F$_\beta$ $\Uparrow$} & \textbf{IOU $\Uparrow$} \\\cmidrule{1-7} \cmidrule{9-13} 
			
			\multirow{3}{*}{\textbf{FCN-finetuned}} & DHS~\cite{Liu-CVPR2016}  & 0.138 & 59.6 & 38.8 & 61.4 & 36.7 &  & 0.072 & 79.6 & 67.9 & 80.8 & 68.1 \\
			& DSS~\cite{Hou-CVPR2017}  & 0.145 & 58.2 & 38.5 & 58.4 & 38.1 &  & 0.076 & 79.0 & 68.6 & 79.2 & 68.7 \\
			& SRM~\cite{Wang-ICCV2017}  & 0.127 & 66.3 & 45.4 & 65.6 & 42.1 &  & 0.067 & 83.0 & 71.7 & 83.1 & 70.8 \\
			& WSS~\cite{Wang-CVPR2017}  & 0.149 & 64.2 & 43.9 & 63.8 & 38.2 &  & 0.085 & 81.1 & 67.8 & 82.0 & 68.7 \\
			
			\midrule
			
			\multirow{3}{*}{\textbf{ANet setting}~\cite{CAMO}} & ANet-DHS & 0.130 & 62.6 & 43.7 & 63.1 & 42.3 &  & 0.072 & 81.2 & 71.2 & 81.4 & 70.5 \\
			& ANet-DSS  & 0.132 & 58.7 & 40.4 & 60.7 & 39.0 &  & 0.067 & 79.5 & 70.1 & 80.4 & 69.4 \\
			& ANet-SRM  & 0.126 & 65.4 & 47.5 & 66.2 & 46.6 &  & 0.069 & 82.6 & 73.2 & 83.0 & 72.7 \\
			& ANet-WSS  & 0.140 & 66.1 & 45.9 & 64.3 & 40.7 &  & 0.078 & 82.6 & 71.0 & 82.0 & 69.7 \\
			
			\midrule
			
			\multirow{3}{*}{\textbf{MirrorNet setting}} 
			& MirrorNet (ResNet-50) & 0.100 & 72.3 & 58.1 & 72.5 & 58.4 &  & 0.062 & 86.1 & 77.9 & 86.2 & 78.0 \\
			& MirrorNet (ResNet-101) & 0.089 & 73.4 & 60.4 & 73.5 & 60.5 &  & 0.053 & 86.7 & 79.4 & 86.7 & 79.4 \\
			
			& MirrorNet (ResNeXt-101) & \textcolor{red}{\textbf{0.084}} & \textcolor{red}{\textbf{76.6}} & \textcolor{red}{\textbf{62.7}} & \textcolor{red}{\textbf{76.9}} & \textcolor{red}{\textbf{62.9}} &  & \textcolor{red}{\textbf{0.051}} & \textcolor{red}{\textbf{88.2}} & \textcolor{red}{\textbf{80.4}} & \textcolor{red}{\textbf{88.4}} & \textcolor{red}{\textbf{80.5}} \\
			& MirrorNet (ResNeXt-152) & \textcolor{blue}{\textbf{0.077}} & \textcolor{blue}{\textbf{78.4}} & \textcolor{blue}{\textbf{65.8}} & \textcolor{blue}{\textbf{78.5}} & \textcolor{blue}{\textbf{65.7}} &  & \textcolor{blue}{\textbf{0.045}} & \textcolor{blue}{\textbf{89.3}} & \textcolor{blue}{\textbf{82.2}} & \textcolor{blue}{\textbf{89.3}} & \textcolor{blue}{\textbf{82.2}} \\
			\bottomrule
		\end{tabular}
	}
\end{table*}

\begin{table*}[t]
	\centering
	\caption{Ablation study results on the full CAMO dataset. The evaluation is based on F-measure~\cite{Achanta-CVPR2009} (the higher the better), IOU~\cite{Long-ICCV2015} (the higher the better), and MAE (the lower the better). The best results are shown in \textcolor{blue}{\textbf{blue}}. The detail of each baseline is clearly introduced in Section~\ref{sec:ablation}.}
	\label{tab:ablations}
	\resizebox{1\textwidth}{!}{
		\begin{tabular}{|l|c|c|c|c|c|c|c|c|c|}
			\toprule
			\multirow{2}{*}{\textbf{Method}} & \multicolumn{4}{c|}{\textbf{Settings}} & \textbf{} & \multicolumn{2}{c|}{\textbf{Adaptive Threshold}} & \multicolumn{2}{c|}{\textbf{Fixed Threshold}} \\
			
			\cmidrule{2-10}
			
			& \textbf{\begin{tabular}[c]{@{}c@{}}Main\\Stream\end{tabular}} & \textbf{\begin{tabular}[c]{@{}c@{}}Mirror\\Stream\end{tabular}} & \textbf{\begin{tabular}[c]{@{}c@{}}Data Augmentation\\in the Wild\end{tabular}} & \textbf{Pooling} & \textbf{MAE $\Downarrow$} & \textbf{F$_\beta$ $\Uparrow$} & \textbf{IOU $\Uparrow$} & \textbf{F$_\beta$ $\Uparrow$} & \textbf{IOU $\Uparrow$}\\
			
			\midrule
			
			Baseline 1 & \checkmark &  & \checkmark & PrRoI & 0.051 & 88.4 & 80.8 & 88.6 & 80.8 \\ \midrule
			Baseline 2 &  & \checkmark & \checkmark & PrRoI & 0.050	& 88.6 &	81.3	& 88.8	& 81.4  \\ \midrule
			Baseline 3 & \checkmark & \checkmark &  & PrRoI & 0.048 & 88.6 & 82.0  & 88.8 & 82.0 \\ \midrule
			Baseline 4 & \checkmark & \checkmark & \checkmark & RoI Align & 0.047 & 88.8 & 81.8 & 88.9 & 81.8 \\ \midrule
			\textcolor{blue}{\textbf{MirrorNet}} & \checkmark & \checkmark & \checkmark & PrRoI & \textcolor{blue}{\textbf{0.045}} & \textcolor{blue}{\textbf{89.3}} & \textcolor{blue}{\textbf{82.2}} & \textcolor{blue}{\textbf{89.4}} & \textcolor{blue}{\textbf{82.2}} \\
			
			\bottomrule
		\end{tabular}
	}
\end{table*}


We conduct the experiments on CAMO dataset with two settings, namely, only camouflaged object set, and the full set. To investigate the impact of different backbones on our proposed MirrorNet, we fine-tuned pre-trained models of ResNet (ResNet-50, ResNet-101)~\cite{He-CVPR2016}, and ResNeXt (ResNeXt-101, ResNeXt-152)~\cite{resnext} on CAMO respectively. 
It is interesting for the computer vision community to see the performance of different frameworks in the camouflaged object segmentation problem. Particularly, we compare our MirrorNet with ANet family networks~\cite{CAMO} (denoted by ANet-DHS, ANet-DSS, ANet-SRM, ANet-WSS), the-state-of-the-art for camouflaged object segmentation, and the original FCNs (DHS~\cite{Liu-CVPR2016}, DSS~\cite{Hou-CVPR2017}, SRM~\cite{Wang-ICCV2017}, and WSS~\cite{Wang-CVPR2017}). All these methods were also fine-tuned on the CAMO dataset with their published parameters.

As shown in Table~\ref{tab:result}, ANet variants outperform FCN-finetuned methods and ANet variants with the same backbone. Meanwhile, our proposed MirrorNet significantly outperforms all baselines. MirrorNet achieves the best result with ResNeXt-152 backbone. This is totally consistent with our discussion in Section~\ref{sec:motivation}, \textit{i.e.}, the better backbones tend to work better in MirrorNet. Note that all baselines, \textit{i.e.}, FCN-finetuned or ANet variants are trained and tested separately on different sets of data, for example, training and testing on only camouflaged image sets, or training and testing on the full set. Meanwhile, our MirrorNet and its variants are only trained on the full set. However, their performance is also significantly better in the only camouflaged image set. 




\subsection{Ablation Study}
\label{sec:ablation}
In this subsection, we investigate the impact of different components in our proposed MirrorNet such as dual stream, data augmentation, and RoI pooling mechanisms. Experimental results are shown in Table~\ref{tab:ablations}. We remark that we use the ResNeXt-152 backbone for all compared methods.

\textbf{Dual Stream: } We investigate the performance of each stream in the MirrorNet, namely, main stream and mirror stream. We compare our completed MirrorNet, which have both streams, with using single streams (denoted by \textit{Baseline 1 for only the main stream} and \textit{Baseline 2 for only the mirror stream}). As shown in Table~\ref{tab:ablations}, our completed MirrorNet outperforms baselines using individual stream. In addition, the mirror stream output is better than the main stream output alone. 

\textbf{Data Augmentation in the Wild: } We suspected that one of the major factors affecting the camouflaged objects segmentation was the limited number of training data. To evaluate the impact of Data Augmentation in the Wild, we compare MirrorNet \textit{without using Data Augmentation in the Wild in the training process, denoted by Baseline 3}. Table~\ref{tab:ablations} shows that using Data Augmentation in the Wild in the training process outperforms the one without using the augmented data. 

\textbf{ROI Pooling: } We further investigate the performance of different ROI pooling mechanism, namely, \textit{RoI Align (denoted by Baseline 4)} and PrROI. As also seen in Table~\ref{tab:ablations}, PrRoI-based MirrorNet surpasses RoI Align-based MirrorNet. In addition, RoI Align-based MirrorNet outperforms the main stream and the mirror stream (with PrRoI). This clearly shows the importance of the mask fusion. 

\begin{figure*}[!t]
	\centering
	\includegraphics[width=0.86\linewidth]{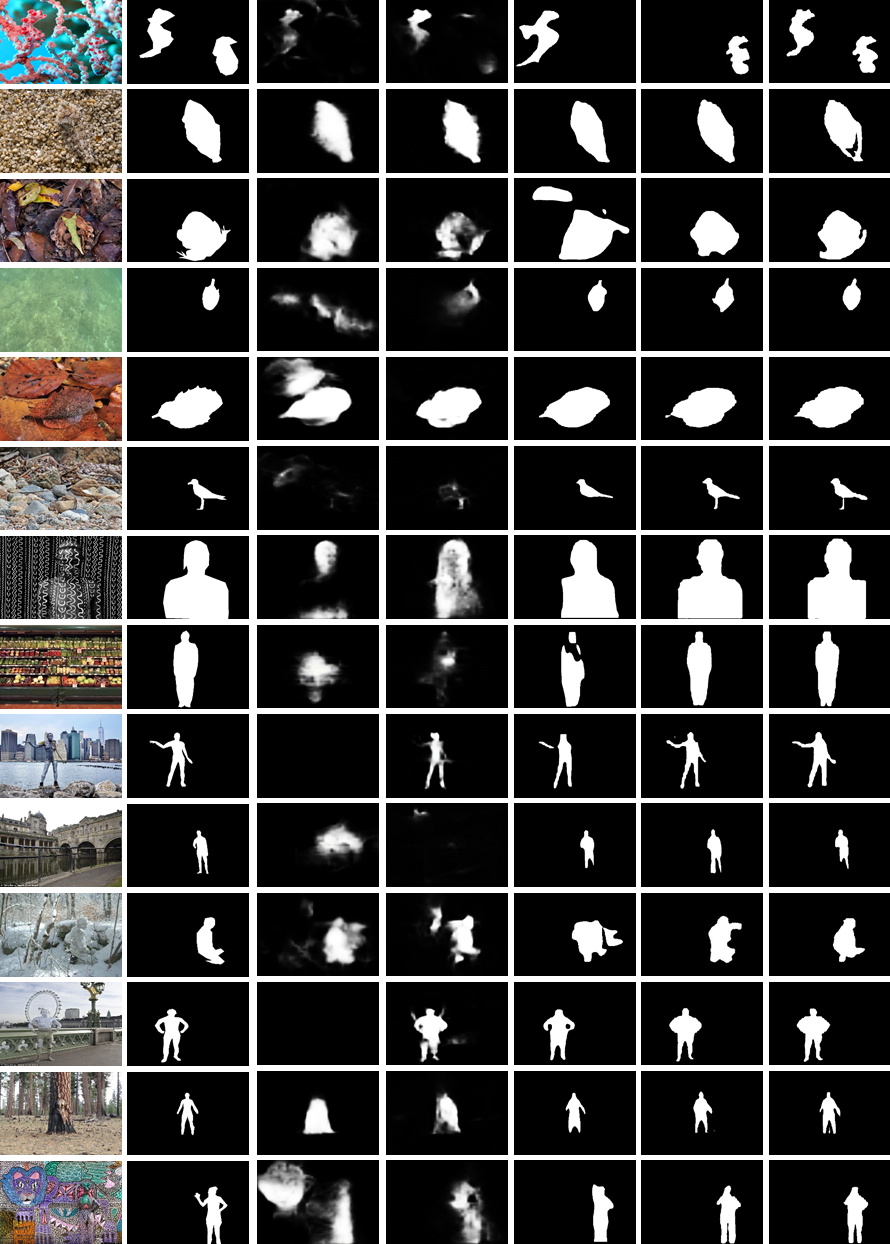}
	\caption{Comparison of the results from different methods. From left to right: the original image, the ground truth map, the predicted camouflaged maps of ANet-SRM~\cite{CAMO}, SINet~\cite{sinet}, and our methods MirrorNet (ResNet-101), MirrorNet (ResNeXt-101), MirrorNet (ResNeXt-152). 
	}
	\label{fig:baselines}
\end{figure*}

\begin{table}[!t]
\centering
	\caption{Comparison of state-of-the-art methods on CAMO dataset (Camouflaged object only). The evaluation is based on  E-measure ($E_{\phi}$)~\cite{ephi} (the higher the better), S-measure ($S_{\alpha}$)~\cite{smeasure} (the higher the better), weighted F-measure ($F_{\beta}^{\omega}$)~\cite{weightedf} (the higher the better), and MAE (the smaller the better). The best results are shown in \textcolor{blue}{\textbf{blue}}.}
\label{tab:sinet}
\footnotesize
\begin{tabular}{|l|l|cccc|}
\toprule
\multirow{2}{*}{\textbf{Method}}
 & \multirow{2}{*}{\textbf{Year}} &
 \multicolumn{4}{c|}{\textbf{Evaluation Metrics}} \\
 \cmidrule{3-6}
 & &
 \multicolumn{1}{c|}{\textbf{S$_{\alpha}$} $\Uparrow$}                            & \multicolumn{1}{c|}{\textbf{E$_{\phi}$} $\Uparrow$}                            & \multicolumn{1}{c|}{\textbf{F$_{\beta}^{\omega}$} $\Uparrow$}                            & \multicolumn{1}{c|}{\textbf{MAE $\Downarrow$}}                            \\ \midrule
\multicolumn{1}{|l|}{UNet++~\cite{unet}}           & \multicolumn{1}{l|}{2018}          & \multicolumn{1}{l|}{0.599}                                 & \multicolumn{1}{l|}{0.653}                                 & \multicolumn{1}{l|}{0.392}                                 & \multicolumn{1}{c|}{0.149}                                 \\ \midrule
\multicolumn{1}{|l|}{PiCANet~\cite{picanet}}          & \multicolumn{1}{l|}{2018}          & \multicolumn{1}{l|}{0.609}                                 & \multicolumn{1}{l|}{0.584}                                 & \multicolumn{1}{l|}{0.356}                                 & \multicolumn{1}{c|}{0.156}                                 \\ \midrule
\multicolumn{1}{|l|}{MSRCNN~\cite{maskscore}}           & \multicolumn{1}{l|}{2019}          & \multicolumn{1}{l|}{0.617}                                 & \multicolumn{1}{l|}{0.669}                                 & \multicolumn{1}{l|}{0.454}                                 & \multicolumn{1}{c|}{0.133}                                 \\ \midrule
\multicolumn{1}{|l|}{PoolNet~\cite{poolnet}}          & \multicolumn{1}{l|}{2019}          & \multicolumn{1}{l|}{0.702}                                 & \multicolumn{1}{l|}{0.698}                                 & \multicolumn{1}{l|}{0.494}                                 & \multicolumn{1}{c|}{0.129}                                 \\ \hline
\multicolumn{1}{|l|}{BASNet~\cite{basnet}}           & \multicolumn{1}{l|}{2019}          & \multicolumn{1}{l|}{0.618}                                 & \multicolumn{1}{l|}{0.661}                                 & \multicolumn{1}{l|}{0.413}                                 & \multicolumn{1}{c|}{0.159}                                 \\ \midrule
\multicolumn{1}{|l|}{PFANet~\cite{pfanet}}           & \multicolumn{1}{l|}{2019}          & \multicolumn{1}{l|}{0.659}                                 & \multicolumn{1}{l|}{0.622}                                 & \multicolumn{1}{l|}{0.391}                                 & \multicolumn{1}{c|}{0.172}                                 \\ \midrule
\multicolumn{1}{|l|}{CPD~\cite{cpd}}              & \multicolumn{1}{l|}{2019}          & \multicolumn{1}{l|}{0.726}                                 & \multicolumn{1}{l|}{0.729}                                 & \multicolumn{1}{l|}{0.550}                                 & \multicolumn{1}{c|}{0.115}                                 \\ \midrule
\multicolumn{1}{|l|}{HTC~\cite{htc}}              & \multicolumn{1}{l|}{2019}          & \multicolumn{1}{l|}{0.476}                                 & \multicolumn{1}{l|}{0.442}                                 & \multicolumn{1}{l|}{0.174}                                 & \multicolumn{1}{c|}{0.172}                                 \\ \midrule
\multicolumn{1}{|l|}{EGNet~\cite{egnet}}            & \multicolumn{1}{l|}{2019}          & \multicolumn{1}{l|}{0.732}                                 & \multicolumn{1}{l|}{0.768}                                 & \multicolumn{1}{l|}{0.583}                                 & \multicolumn{1}{c|}{0.104}                                 \\ \midrule
\multicolumn{1}{|l|}{ANet-SRM~\cite{CAMO}}         & \multicolumn{1}{l|}{2019}          & \multicolumn{1}{l|}{0.682}                                 & \multicolumn{1}{l|}{0.685}                                 & \multicolumn{1}{l|}{0.484}                                 & \multicolumn{1}{c|}{0.126}                                 \\ \midrule
\multicolumn{1}{|l|}{SINet~\cite{sinet}}            & \multicolumn{1}{l|}{2020}          & \multicolumn{1}{l|}{0.751}                                 & \multicolumn{1}{l|}{0.771}                                 & \multicolumn{1}{l|}{0.606}                                 & \multicolumn{1}{c|}{0.100}                                 \\ \midrule
\multicolumn{1}{|l|}{\textbf{MirrorNet (Ours)}}        & \multicolumn{1}{l|}{\textbf{2020}}          & \multicolumn{1}{l|}{{\color{blue} \textbf{0.785}}} & \multicolumn{1}{l|}{{\color{blue} \textbf{0.849}}} & \multicolumn{1}{l|}{{\color{blue} \textbf{0.719}}} & \multicolumn{1}{c|}{{\color{blue} \textbf{0.077}}} \\ \bottomrule
             
\end{tabular}
\end{table}

\subsection{Comparison with State-of-the-art Methods}
We further compare the performance of MirrorNet with state-of-the-arts reported in~\cite{sinet}. Table~\ref{tab:sinet} shows the performance of different methods in terms of E-measure ($E_{\phi}$)~\cite{ephi},  S-measure ($S_{\alpha}$)~\cite{smeasure}, weighted F-measure ($F_{\beta}^{\omega}$)~\cite{weightedf}, and MAE~\cite{CAMO}. Note that these metrics were proposed in~\cite{sinet}. As can be seen, the recently proposed methods tend to achieve better results. Our proposed method, MirrorNet, achieves the best performance in terms of $E_{\phi}$,  $S_{\alpha}$, $F_{\beta}^{\omega}$, and MAE. MirrorNet with ResNeXt-152 backbone surpasses the state-of-the-art methods by a remarkable margin in all metrics. 


Figure \ref{fig:baselines} shows the visual comparison of different methods. As illustrated in the figure, our MirrorNet variants yield better results than all state-of-the-art methods. Our results are close to the ground truth and focus on the camouflaged objects. From the results, our MirrorNet successfully segments the camouflaged objects images in CAMO dataset~\cite{CAMO}. As discussed in~\cite{CAMO}, CAMO dataset consists of images from challenging scenarios such as object appearance (the camouflaged object has similar color and texture with the background), background clutter, shape complexity, small object, object occlusion, multiple objects, and distraction. This demonstrates that our MirrorNet can well handle a variety of camouflaged objects. 

\section{Conclusion}
\label{sec:conclusion}

\hlcyan{In this paper, we propose a simple and flexible end-to-end network, namely MirrorNet, for camouflaged object segmentation. Our bio-inspired MirrorNet leverages both instance segmentation and mirror stream to segment camouflaged objects in images. Particularly, we propose to use the main  stream and the mirror stream, which is embedded image flipping, to effectively capture different layouts of the scene, leading to significant performance in identifying camouflaged objects. The extensive experimental results show that our proposed method achieves state-of-the-art performance on the public CAMO dataset.}

\hlcyan{In the future, we will investigate different adversarial schemes. In addition, we aim to further explore the problem of camouflaged instance segmentation. Through experiments, our data augmentation slightly improves the performance as shown in the ablation study. Note that data augmentation is not the main focus of this paper. Exploring different data augmentation methods is an interesting topic which is left for future work. We are also interested in applying camouflaged object segmentation into practical systems}~\cite{TamIJCV}.

\section*{Acknowledgment}

This research is in part granted by Vingroup Innovation Foundation (VINIF) in project code VINIF.2019.DA19, National Science Foundation (NSF) under Grant No. 2025234, and JSPS KAKENHI Grant No. 20K23355. We gratefully acknowledge NVIDIA.

\ifCLASSOPTIONcaptionsoff
  \newpage
\fi

 \balance

\bibliography{short_bibtex}
\bibliographystyle{IEEEtran}









\vfill 




\end{document}